\documentclass[final]{cvpr}
\usepackage{times}
\usepackage{epsfig}
\usepackage{graphicx}
\usepackage{amsmath}
\usepackage{amssymb}
\usepackage{enumitem}
\usepackage{algpseudocode}
\usepackage{dsfont}
\usepackage{multirow}
\usepackage{booktabs}
\usepackage{xcolor}
\definecolor{linkc}{rgb}{0, 0.44, 0.74}
\definecolor{eqc}{rgb}{1, 0, 0}

\usepackage[pagebackref=false,breaklinks=true,letterpaper=true,colorlinks,urlcolor=linkc,citecolor=linkc,linkcolor=eqc,bookmarks=false]{hyperref}

\newcommand{\figdir}{figures}

\def\cvprPaperID{5184} 
\def\confYear{CVPR 2021}

\begin{document}

\title{VideoMoCo: Contrastive Video Representation Learning with \\ Temporally Adversarial Examples}

\author{
Tian Pan$^1$\thanks{T. Pan and Y. Song contribute equally. Y. Song is the corresponding author. The code is available at \url{https://github.com/tinapan-pt/VideoMoCo}.}\and
Yibing Song$^{1\ast}$ \and
Tianyu Yang$^1$ \and
Wenhao Jiang$^2$ \and
Wei Liu$^2$ \and
$^1$Tencent AI Lab \quad $^2$Tencent Data Platform \\
{\tt\small pantianauraro@foxmail.com\quad yibingsong.cv@gmail.com\quad tianyu-yang@outlook.com}\\
{\tt\small cswhjiang@gmail.com\qquad wl2223@columbia.edu}
}

\maketitle

\begin{abstract}
MoCo~\cite{he2020momentum} is effective for unsupervised image representation learning. In this paper, we propose VideoMoCo for unsupervised video representation learning. Given a video sequence as an input sample, we improve the temporal feature representations of MoCo from two perspectives. First, we introduce a generator to drop out several frames from this sample temporally. The discriminator is then learned to encode similar feature representations regardless of frame removals. By adaptively dropping out different frames during training iterations of adversarial learning, we augment this input sample to train a temporally robust encoder. Second, we use temporal decay to model key attenuation in the memory queue when computing the contrastive loss. As the momentum encoder updates after keys enqueue, the representation ability of these keys degrades when we use the current input sample for contrastive learning. This degradation is reflected via temporal decay to attend the input sample to recent keys in the queue. As a result, we adapt MoCo to learn video representations without empirically designing pretext tasks. By empowering the temporal robustness of the encoder and modeling the temporal decay of the keys, our VideoMoCo improves MoCo temporally based on contrastive learning. Experiments on benchmark datasets including UCF101 and HMDB51 show that VideoMoCo stands as a state-of-the-art video representation learning method.

\end{abstract}

\begin{figure}
    \centering
    \begin{tabular}{c}
        \includegraphics[width=0.95\linewidth]{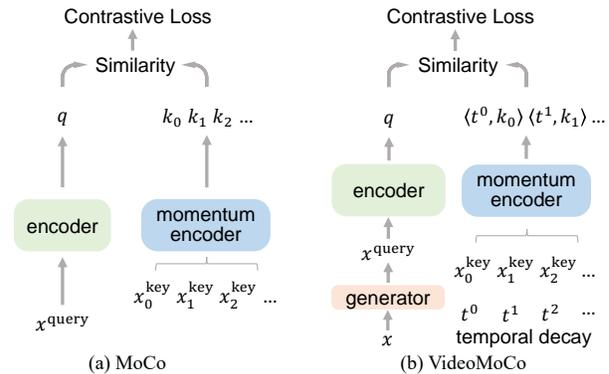}\\
    \end{tabular}
    \vspace{1mm}
    \caption{VideoMoCo improves MoCo~\cite{he2020momentum} temporally from two perspectives. First, by taking a video sequence as a training sample, we introduce adversarial learning to augment this sample temporally. Second, we use a temporal decay (i.e., $t^i$) to attenuate the contributions from older keys in the queue. To this end, the encoder is learned via temporal augmentation within each sample and temporally contrastive learning across different samples.}
    \label{fig:teaser}
\end{figure}

\section{Introduction}
Unsupervised (i.e., self-supervised) feature representation learning receives tremendous investigations along with the development of convolutional neural networks (CNNs). This learning scheme does not require cumbersome manual label collections and produces deep features representing general visual contents. These features can be further adapted to suit downstream visual recognition scenarios including image classification~\cite{kolesnikov2019revisiting,zhan2020online}, object detection~\cite{wu2018unsupervised,kolesnikov2019big}, visual tracking~\cite{wang2019unsupervised,wang2020unsupervised}, and semantic segmentation~\cite{sabokrou2019self}. Among various unsupervised representation learning studies, contrastive learning~\cite{hadsell2006dimensionality} is developed extensively. By treating one sample as positive and the remaining ones as negative (i.e., instance discrimination), contrastive learning improves feature discrimination by considering the massive sample storage~\cite{wu2018unsupervised,he2020momentum,tian2020contrastive,misra2020self} and data augmentation~\cite{chen2020simple}. To effectively handle large-scale samples, MoCo~\cite{he2020momentum} builds an on-the-fly dictionary with a queue and a moving-averaged encoder. The learned feature representations have significantly improved a series of downstream recognition performances to approach those by using supervised feature representations.

The evolution of contrastive learning heavily focuses on feature representations from static images while leaving the temporal video representations less touched. One of the reasons is that large-scale video data is difficult to store in memory. Using a limited number of video samples leads to inadequate contrastive learning performance. On the other hand, attempts on unsupervised video representation learning focus on proposing pretext tasks related to a sub-property of video content. Examples include sequence sorting~\cite{lee2017unsupervised}, optical flow estimation~\cite{gan2018geometry}, video playback rate perception~\cite{yao2020video}, pace prediction~\cite{wang2020self}, and temporal transformation recognition~\cite{jenni2020video}. Different from these pretext designs, we aim to learn a task-agnostic feature representation for videos. With effective data storage and CNN update~\cite{he2020momentum} at hand, we rethink unsupervised video representation learning from the perspective of contrastive learning, where the features are learned naturally to discriminate different video sequences without introducing empirically designed pretext tasks.

In this work, we propose VideoMoCo that improves MoCo for unsupervised video representation learning. VideoMoCo follows the usage of queue structure and a moving-averaged encoder of MoCo, which computes a contrastive loss (i.e., InfoNCE~\cite{oord2018representation}) efficiently among large-scale video samples. Given a training sample with fixed-length video frames, we introduce adversarial learning to improve the temporal robustness of the encoder. As shown in Fig.~\ref{fig:pipelineG}, we use a generator (G) to adaptively drop out several frames. The sample with remaining frames, together with the original sample with full frames, are sent to the discriminator / encoder (D) for differentiating. Their difference is then utilized reversely to train G. As a result, G removes temporally important frames based on the current state of D. And D is learned to produce similar feature representations regardless of frame removal. During different training iterations, the frames removed by G are different. This sample is then augmented adversarially to train a temporally robust D. After adversarial learning only D is kept to extract temporally robust features.

The adversarial learning drops out several frames of an input video sample. We treat its remaining frames as a query sample and perform contrastive learning with keys in the memory queue. However, we notice that the momentum encoder updates after keys enqueue. The feature representation of the keys is not up-to-date when we compute the contrastive loss. To mitigate this effect, we model the degradation of these keys by proposing a temporal decay. If a key stays longer in the queue, its contribution is less via this decay. To this end, we attend the query sample to recent keys during the contrastive loss computation. The encoder is thus learned more effectively without being heavily interfered by the 'ancient' keys. The temporally adversarial learning and the temporal decay improve the temporal feature representation of MoCo. The experiments on several action recognition benchmarks verify that our proposed VideoMoCo performs favorably against state-of-the-art video representation approaches.

We summarize our main contributions as follows:
\begin{itemize}[noitemsep,nolistsep]
  \item We propose temporally adversarial learning to improve the feature representation of the encoder.
  \item We propose a temporal decay to reduce the effect from historical keys in the memory queues during contrastive learning.
  \item Experiments on the standard benchmarks show that our VideoMoCo extends MoCo to a state-of-the-art video representation learning method.
\end{itemize}

\section{Related Work}
In this section, we perform a literature review on
self-supervised video representation learning, contrastive learning, and generative adversarial learning.

\subsection{Video Representation Learning}
Investigations on video representation learning focus on exploiting temporal coherence among consecutive video frames. In~\cite{dwibedi2019temporal,wang2019learning}, cycle consistency is explored for feature learning. The future frame prediction is proposed in~\cite{srivastava2015unsupervised} to learn coherent features. A set of studies propose empirical and unsupervised feature learning pretext tasks including future motion and appearance prediction~\cite{wang2019self}, video pace prediction~\cite{wang2020self}, and frame color estimation~\cite{vondrick2018tracking}. In addition to future frame prediction, video frame sorting~\cite{lee2017unsupervised,xu2019self,kim2019self,fernando2017self} is popular. In~\cite{he2019bag}, contrastive loss is applied to learn video representations by comparing frames from multiple viewpoints. A consistent frame representation of different sampling rates is proposed in~\cite{yang2020video}. Four different temporal transformations (i.e., speed change, random sampling, periodic change, and content warp) of a video are investigated in~\cite{jenni2020video} to build video representations for action recognition. In addition, multi-modality methods introduce text~\cite{sun2019videobert}, optical flows~\cite{gan2018geometry}, and audios~\cite{korbar2018cooperative,wang2019learning,dwibedi2019temporal} for cross-modal supervised learning. Different from existing video representation learning methods, our VideoMoCo utilizes color information and performs instance discrimination without bringing empirical pretext tasks.

\subsection{Contrastive Learning}
There are wide investigations in contrastive learning for static image recognition. They follow the principle that the feature distances between positive sample pairs are minimized while those between the negative sample pairs are maximized during the training stage. The distance measurement is proposed in~\cite{hadsell2006dimensionality}. The instance discrimination is proposed in~\cite{wu2018unsupervised} for feature learning. This scheme is improved in MoCo~\cite{he2020momentum} where there is a momentum encoder to build dynamic dictionaries to support contrastive learning. In SimCLR~\cite{chen2020simple}, different combinations of data augmentation methods are evaluated for sample pairs. Only positive samples are introduced in BYOL~\cite{grill2020bootstrap} during training while achieving superior performance. Its effectiveness is justified by the theoretical analysis~\cite{poole2019variational}. The objectives utilized in contrastive learning are to maximize the lower bound of mutual information between the input feature and its augmented representation~\cite{linsker1988self}. Different from existing methods that focus on static image representations, we improve MoCo temporally to gain robust video representations.

\subsection{Generative Adversarial Learning}
The adversarial learning is developed in~\cite{goodfellow2014generative} where CNN is introduced to generate realistic image appearances from random noises. There are two subnetworks in a typical generative adversarial network (GAN). One is the generator and the other is the discriminator. The goal of the generator is to synthesize images that can fool the discriminator, while the discriminator is learned to distinguish between the real images and the synthetic ones from the generator. The generator and the discriminator are trained simultaneously by competing with each other. This unsupervised training mechanism outperforms traditional supervised training scheme to produce realistic image content. There are tremendous investigations on analyzing the GAN training~\cite{martin2017wasserstein,nowozin2016f,gurumurthy2017deligan}. Meanwhile, there are many computer vision applications of GAN including image generation~\cite{isola2017image,liu2020rethinking,wang2020rethinking}, object tracking~\cite{song2018vital,song2017crest,jia2020robust}, and semantic segmentation~\cite{souly2017semi}. Different from major GAN methods for generator learning, we augment training data in an adversarial form to train a temporally robust discriminator. This learning process is specially integrated into contrastive learning. The discriminator is our intended encoder to capture video representations.

\begin{figure*}[t]
    \centering
    \begin{tabular}{c}
        \includegraphics[width=0.9\linewidth]{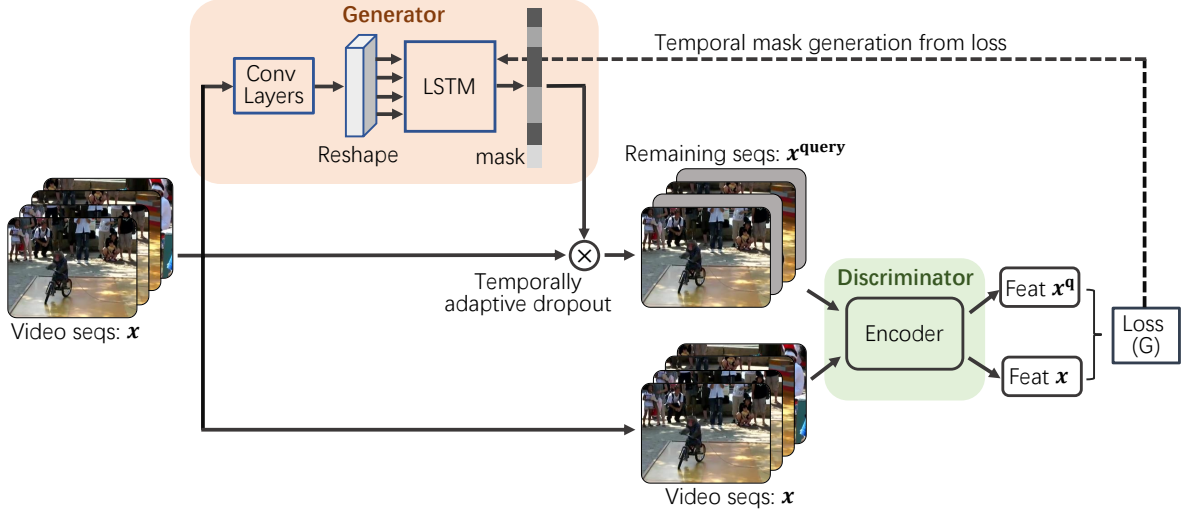}\\
    \end{tabular}
    \caption{Temporally adversarial learning. The input sample $x$ is a video clip. The generator G containing LSTM dropouts several frames of $x$ to generate $x^{\rm query}$. The discriminator D (i.e., encoder) extracts features from both $x$ and $x^{\rm query}$ and compute their similarity loss. We use this loss term reversely to train G. During training iterations, G is learned to continuously attack D by removing different frames of $x$ (i.e., temporal data augmentation), and D is learned to defend this attack by encoding temporally robust features. We use $x^{\rm query}$ to perform contrastive learning to train D as shown in Fig.~\ref{fig:pipelineD}.}
    \label{fig:pipelineG}
\end{figure*}

\begin{figure}
    \centering
    \begin{tabular}{c}
        \includegraphics[width=0.9\linewidth]{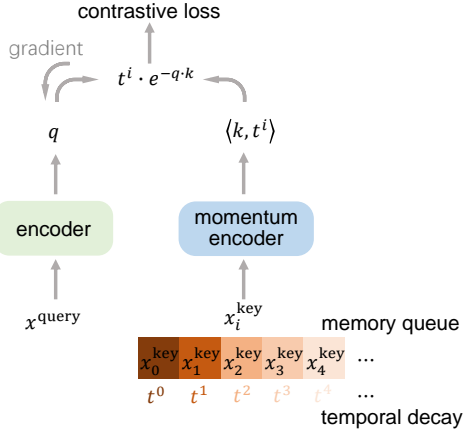}\\
    \end{tabular}
    \vspace{1mm}
    \caption{Temporally contrastive learning. We follow MoCo to use a memory queue to store keys. Besides, we introduce a temporal decay $t^i$ ($t\in(0,1)$) to model the attenuation of each key in this queue during contrastive learning.}
    \label{fig:pipelineD}
\end{figure}

\section{VideoMoCo}\label{sec:method}
Our VideoMoCo is built upon MoCo for unsupervised video representation learning. We first briefly review MoCo and then present our temporally adversarial learning and temporal decay.
Furthermore, we visualize the temporal robustness of the feature by showing the entropy values of the classifier and the network attentions.

\subsection{MoCo Overview}\label{sec:moco}
Momentum Contrast (MoCo) provides a dictionary look-up for contrastive learning. Given an encoded query $q$ and encoded keys $\{k_0,k_1,k_2,...\}$ in a dictionary queue, the contrastive loss of MoCo can be written as:
\begin{equation}\label{eq:nce}
    L_q=-\operatorname{log}\frac{\operatorname{exp}(q\cdot k_+/\tau)}{\sum_{i=0}^K \operatorname{exp}(q\cdot k_i/\tau)}
\end{equation}
where $\tau$ is a scalar. The sum is over one positive and $K$ negative samples. This loss tends to classify $q$ as $k_+$ via a softmax classification process. The query $q$ is the representation of an input sample via the encoder network, while the keys $k_i$ are the representations of the other training samples in the queue.

The core of momentum contrast is to dynamically maintain the queue. The samples in the queue are progressively replaced following an FIFO (first in, first out) scheme. After computing the contrastive loss in Eq.~\eqref{eq:nce}, the encoder is updated via gradients while the momentum encoder is updated as a moving-average of the encoder weights. We denote the parameters of an encoder as $\theta_q$ and those of a momentum encoder as $\theta_k$. The momentum encoder is updated as:
\begin{equation}\label{eq:moco}
\theta_k \leftarrow m\theta_k+(1-m)\theta_q
\end{equation}
where $m\in[0,1)$ is a momentum coefficient. The momentum encoder is updated slowly based on the encoder change, which ensures stable key representations.

VideoMoCo improves MoCo by introducing temporally adversarial learning and temporal decay. Given an input video clip with a fixed number of frames, we send it to a generator and the encoder to produce $q$. Meanwhile, we reweigh $\operatorname{exp}(q\cdot k_i/\tau)$ by using $t^i$ where $t\in(0,1)$. Then, we follow MoCo to train the encoder and update the momentum encoder accordingly. During inference, we remove G and only use the encoder for feature extraction.

\subsection{Temporally Adversarial Learning}
\label{sec:gan}
We propose adversarial learning as a temporal data augmentation strategy to improve feature representations. Fig.~\ref{fig:pipelineG} shows an overview. We have an input sample $x$ where there are a fixed number of frames. The generator G takes $x$ as input and produces a temporal mask. The architecture of G follows ConvLSTM~\cite{xingjian2015convolutional}. The output of ConvLSTM predicts the importance of each frame. We drop out  $25\%$ of the frames with high importance values by using the temporal mask. We denote the output of G as $G(x)$. Then the query sample $x^{\rm query}$ can be written as:
\begin{equation}\label{eq:query}
    x^{\rm query}=G(x)\otimes x
\end{equation}
where $\otimes$ indicates the temporal dropout operation. The feature map size of $x^{\rm query}$ is the same as that of $x$ while the content of the $25\%$ of its frames is removed.

We regard the encoder of MoCo as the discriminator. After obtaining $x^{\rm query}$, we send both $x^{\rm query}$ and $x$ to the discriminator D (i.e., the encoder). The feature representations of $D(x^{\rm query})$ and $D(x)$ are expected to become similar. We use $L_1$-norm as the loss function to train G, which can be written as:
\begin{equation}\label{eq:adv_g}
    \mathop{max}\limits_{G} L(G)=\mathbb{E}_{x\sim P_x} (\left| D(x^{\rm query})-D(x)\right|_1)
\end{equation}
where $P_x$ is the data distribution of $x$. When training $D$, we use the contrastive loss akin to Eq.~\eqref{eq:nce}, which can be written in the form of adversarial loss as:
\begin{equation}\label{eq:adv_d}
\mathop{min}\limits_{D} L(D)=-\operatorname{log}
\frac{\operatorname{exp}(D(x^{\rm query})\cdot k_+/\tau)}
{\sum_{i=0}^K\operatorname{exp}(D(x^{\rm query})\cdot k_i/\tau)}
\end{equation}
where the discriminator / encoder is learned to encode temporally robust feature representations.

We take turns to train G and D during each iteration of adversarial learning. After training D, we update the momentum encoder by using Eq.~\eqref{eq:moco}. Initially, we train D without using G and only use contrastive learning shown in Eq.~\eqref{eq:nce}. When D is learned to approach a semi-stable state, we train D via Eq.~\eqref{eq:adv_d} by involving G. We empirically observe that utilizing adversarial learning at the initial stage makes D difficult to converge.

\subsection{Temporal Decay}\label{sec:decay}
The adversarial learning illustrated in Sec.~\ref{sec:gan} operates on a mini-batch input following MoCo. When computing the contrastive loss, we notice that MoCo treats the contribution of keys from the queue only based on their representations as shown in Eq.~\eqref{eq:nce}. In practice, the momentum encoder evolves after keys entering the queue. The longer the keys in the queue, the more different their representations are compared to those of the current input samples.

In order to mitigate the discrepancy brought by momentum encoder evolvement, we propose a temporal decay to model key degradations. For each key $k_i$ in the queue, we set its corresponding temporal decay as $t^i$ where $t\in(0,1)$. Note that this key moves towards the end of the queue during each training iteration. Thus $i$ gradually increases by 1 and $t^i$ decreases correspondingly. We can rewrite Eq.~\eqref{eq:adv_d} by involving the temporal decay as:
\begin{equation}\label{eq:temp_d}
\mathop{min}\limits_{D} L(D)=-\operatorname{log}
\frac{\operatorname{exp}(D(x^{\rm query})\cdot k_+/\tau)}
{\sum_{i=0}^K t^i\cdot \operatorname{exp}(D(x^{\rm query})\cdot k_i/\tau)}
\end{equation}
where keys contribute to the current sample differently according to their existence time in the queue. In practice, we use Eq.~\eqref{eq:temp_d} and Eq.~\eqref{eq:adv_g} alternatively to train G and D. The remaining training procedure follows those of MoCo.

\begin{figure*}[t]
    \centering
    \begin{tabular}{c}
        \includegraphics[width=1.0\linewidth]{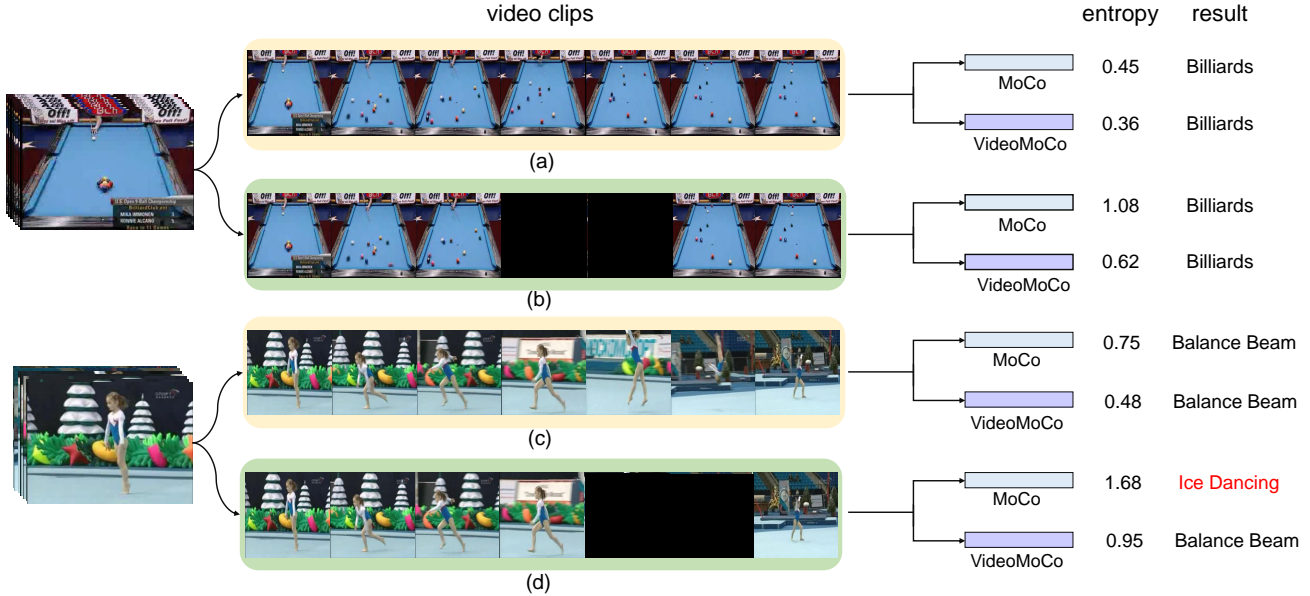}\\
    \end{tabular}
    \caption{Visualizations of CNN predictions between MoCo and VideoMoCo. Given a video sequence, we compute the entropy of the CNN predictions and the classification results. A higher entropy value indicates the CNN classifier is more uncertain about the current predictions. Although both MoCo and VideoMoCo classify video sequences correctly as shown in (a) and (c), the entropy of VideoMoCo is lower than that of MoCo. Meanwhile, if video sequences are partially occluded as shown in (b) and (d), the entropy of VideoMoCo does not increase as much as that of MoCo. The temporally robust feature representations empower VideoMoCo to make correct predictions.}
    \label{fig:vis}
\end{figure*}

\subsection{Visualization}\label{sec:vis}
VideMoCo improves the temporal feature representation of MoCo by incorporating temporally adversarial learning and temporal decay. In this section, we show some visualizations on how temporal feature representations improve the video classification process. Given an input video sequence, MoCo and VideoMoCo both produce a set of classification scores. We use entropy to measure the classifier's confidence when making the prediction. The entropy computation can be written as follows:
\begin{equation}\label{eq:entropy}
H = -\sum_{i=1}^N p(x_i) \operatorname{log}p(x_i)
\end{equation}
where $p(x_{i})$ is the predicted probability of the $i$-th category. When the entropy value is high, it indicates that the classifier is more uncertain on making the current predictions.

We show two input video sequences \textit{Billiards} and \textit{BalanceBeam} in Fig.~\ref{fig:vis}. The complete \textit{Billiards} video is shown in (a), both MoCo and VideoMoCo predict correctly. When we compute the entropy values of these two methods, we observe that VideoMoCo achieves a lower value than that of MoCo. This indicates that although both VideoMoCo and MoCo accurately predict the input video sequence, VideoMoCo is more confident about the current prediction. On the other hand, we increase the difficulty of (a) by partially occluding several frames as shown in (b). To this end, the entropy value of MoCo increases significantly while the value of VideoMoCo does not. This indicates that even though both MoCo and VideoMoCo can resist temporally occluded video sequences, the feature representation of MoCo has become very fragile while that of VideoMoCo does not degrade significantly.

Besides the \textit{Billiards} video sequence, we show the \textit{BalanceBeam} sequence in (c) where these two methods predict correctly. We notice that the entropy values of MoCo and VideoMoCo in (c) are much higher than those in (a). This indicates that compared to the \textit{Billiards} sequence, the \textit{BalanceBeam} sequence is more challenging. When we partially occlude several frames in (d), MoCo predicts incorrectly as \textit{IceDancing}. In contrast, VideoMoCo predicts well. This accurate prediction indicates that the feature representation of VideoMoCo is temporally robust. The entropy values computed based on these two sequences indicate that feature representations in VideoMoCo are more robust than those of MoCo in the temporal domain.

\renewcommand{\tabcolsep}{1pt}
\def\swthree{0.3\linewidth}
\begin{figure}
    \centering
    \small
    \begin{tabular}{ccc}
        \includegraphics[width=\swthree]{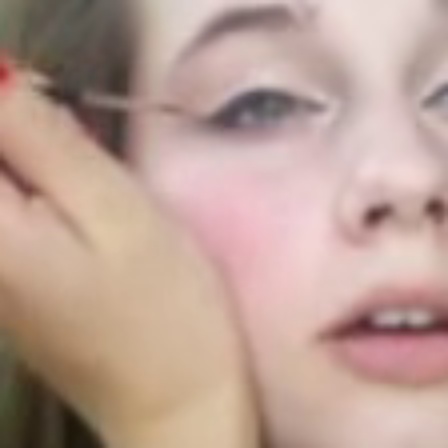}&
        \includegraphics[width=\swthree]{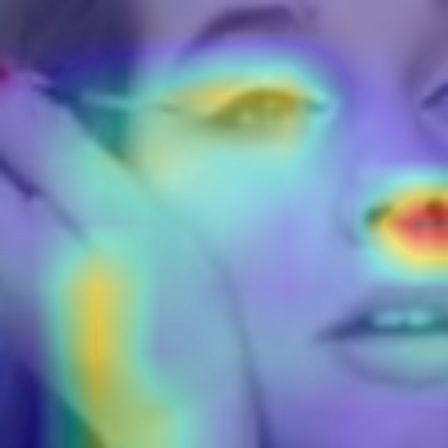}&
        \includegraphics[width=\swthree]{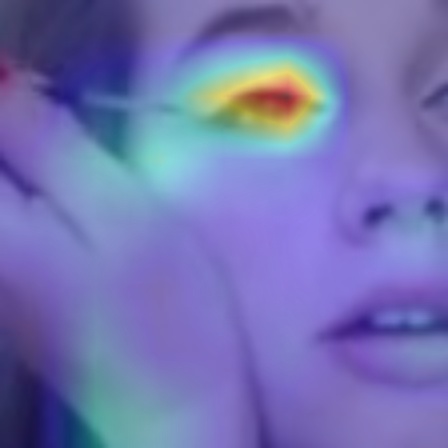}\\
        \includegraphics[width=\swthree]{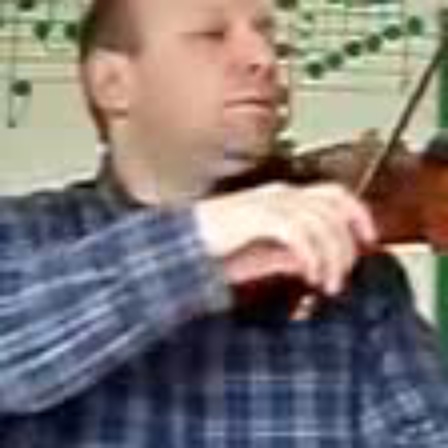}&
        \includegraphics[width=\swthree]{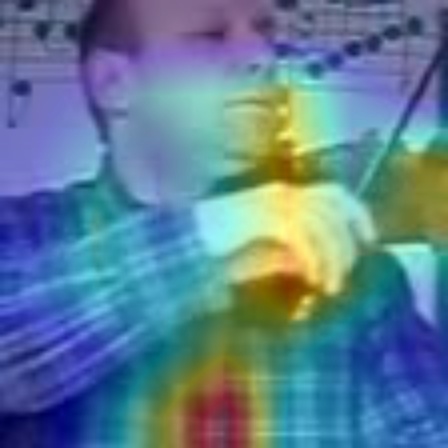}&
        \includegraphics[width=\swthree]{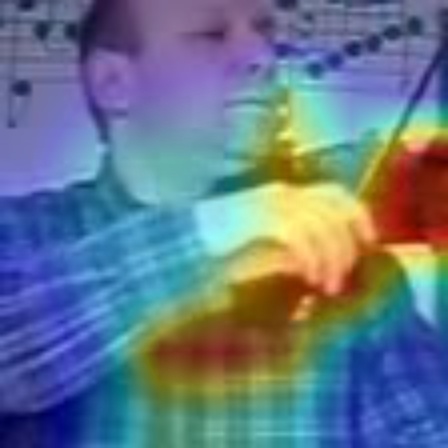}\\
        (a) Video frame&(b) MoCo&(c) VideoMoCo\\
    \end{tabular}
    \vspace{2mm}
    \caption{Attention visualization of MoCo and VideoMoCo. We show video frames in (a), the attention maps from MoCo in (b), and the attention maps from VideoMoCo in (c). In the attention visualization maps, pixels marked as red indicate that the network pays more attention to the current region.}
    \label{fig:map}
\end{figure}

Besides entropy computation, we compute the attention maps of MoCo and VideoMoCo to visualize network attentions. Fig.~\ref{fig:map} shows the visualization result. Two input frames are in (a). The attention maps from MoCo are in (b), and the attention maps from VideoMoCo are in (c). We note that on the first row, VideoMoCo focuses more on the region where eyebrow pencil tip largely moves. This indicates that the features learned from VideoMoCo attend the network to the temporal motions. In comparison, the attention map of MoCo shows that the network does not pay much attention. In the second row, the attention map from VideoMoCo contains higher attention values at the hand region than that of MoCo. The attention map visualization indicates that VideoMoCo is more effective in concentrating on the video motion areas.

\section{Experiments}
In this section, we introduce implementation details and training configurations of VideoMoCo on the K-400 dataset. To analyze the positive effects brought by the temporally adversarial learning and temporal decay, we conduct an ablation study to show the performance improvement by using each of them. Also, we compare VideoMoCo with state-of-the-art self-supervised video representation methods on two standard action recognition benchmarks UCF101 and HMDB51.

\subsection{Benchmark Datasets}
We illustrate the details of benchmark datasets used for the training and inference stages.

{\flushleft \bf Kinetics-400 Dataset.}
Kinetics-400 (K-400~\cite{kay2017kinetics}) is a large-scale action recognition dataset. There are 400 human action categories in total and the whole video sequence amount is 306k. This dataset is split as training, validation, and inference parts. We use the video sequences of the training part to train VideoMoCo. The number of training video sequences is about 240k.

{\flushleft \bf UCF101 Dataset.}
The UCF101 dataset~\cite{soomro2012dataset} is widely used for action recognition. This video sequence is collected from the Internet with predefined 101 action categories. There are over 13k video clips consuming 27 hours. The whole dataset is divided into three training and testing splits. We use the training split 1 to finetune the feature backbone and use the testing split 1 for performance evaluation.

{\flushleft \bf HMDB51 Dataset.}
There are 101 video clips in the HMDB51 dataset~\cite{jhuang2011large} with 51 action categories. This dataset is divided into three training and testing splits. We use the training split 1 to finetune the feature backbone and use the testing split 1 for performance evaluation.

\subsection{Implementation Details}
We illustrate the details of how to integrate the generator G into the pretext contrastive learning process. The discriminator D is the encoder to be utilized for feature extraction. After contrastive learning, we remove G and only use the encoder for downstream finetuning. The downstream finetuning is the same as existing unsupervised video representation learning methods.

{\flushleft \bf Pretext Contrastive Learning.}
Our generator is based on ConvLSTM~\cite{xingjian2015convolutional} where there is one LSTM with 256 hidden units for temporal feature extraction. After taking input video sequences, ConvLSTM first uses several convolutional layers to extract features and reshapes these features into vector forms. The vector is then sent to the LSTM to predict the temporal importance of each video frame. We follow prior work to utilize the small variant of the C3D architecture as our encoder. When training on the K-400 dataset, we consecutively sample 32 frames as a training sample. Note that we sample 32 frames densely within one video sequence to create multiple training samples. After obtaining these training samples, we perform a randomly spatial crop for a fixed size of the input (i.e., $32\times112\times112\times3$). Meanwhile, we apply horizontal flipping, color jittering, and random decolorization with each frame. During the training process, we send each sample into G and drop out $k$ frames with high importance values. Then, this sample is regarded as the query to train D as illustrated in Eq.~\ref{eq:query}. We use an SGD solver to iteratively optimize both G and D with an initial learning rate of 0.02. The momentum is set as 0.9 and the batch size is 128. In practice, we first take 100 epochs to train D for initialization, and then train G and D via adversarial learning for the remaining 100 epochs. We find this empirical design effective to train the encoder.

{\flushleft \bf Downstream finetuning.}
We use two datasets, UCF101 and HMDB51, to finetune the learned feature backbone. The discriminator is kept after pretext training and we regard it as the encoder. The weights of the fully-connected layer are randomly initialized during the finetuning stage. On each dataset, we train the whole network for 10 epochs and then evaluate their performance on the testing data. The experimental setups and data pre-processing method are the same as those of the pretext stage except that the batch size is set as 32 and the learning rate is 0.05.

{\flushleft \bf Evaluation.}
We adopt the standard evaluation protocol~\cite{xu2019self,luo2020video} during testing. To predict each video sequence, we take 10 video clips uniformly from each testing video sequence and average these prediction results. The top-1 accuracy in the action recognition metric is utilized to measure recognition accuracies.

\subsection{Ablation Studies}

We improve MoCo by integrating temporally adversarial learning and temporal decay. In this section, we analyze how these two modules improve the original performance of MoCo. The analysis is based on the configuration that we first train MoCo via pretext and downstream finetuning by using video data. Then, we integrate temporally adversarial learning and temporal decay independently into MoCo during the pretext training. Finally, we combine both of them within MoCo during pretext training. The downstream finetuning is the same for all the configurations. The feature backbone we use is R(2+1)D. UCF101 is used as the training and testing dataset.

\begin{table}
\begin{center}
\caption{Ablation analysis on adversarial learning and dropout amount $k$ on UCF101 and HMDB51 datasets. We set $k=8$ for the random dropout. Adversarial learning improves the baseline while random dropout decreases. $k=8$ is optimal to achieve balanced adversarial learning.}
\label{tab:adversararial}
\vspace{1mm}
\renewcommand{\tabcolsep}{0.9mm}
\begin{tabular}{cccccc}
\toprule
\multicolumn{4}{c}{Experimental Setup} & \multicolumn{2}{c}{Downstream Tasks}\\
\hline
Decay & Adv. & Network & Method & UCF101& HMDB51\\
\hline
$\times$ &$\times$ & R(2+1)D & baseline & 75.9 & 46.2\\
$\times$ &$\times$ & R(2+1)D & random & 74.1 & 43.6 \\
\hline
$\times$ &$\checkmark$ & R(2+1)D & $k=4$ & 76.4 & 48.6\\

$\times$ &$\checkmark$ & R(2+1)D & $k=8$ & 77.8 & 49.1\\

$\times$ &$\checkmark$ & R(2+1)D & $k=12$ & 77.1 & 45.6\\

$\times$ &$\checkmark$ & R(2+1)D & $k=16$ & 75.2 & 40.3\\

$\times$ &$\checkmark$ & R(2+1)D & $k=24$ & 73.4 & 39.0 \\
\hline
$\times$ &$\checkmark$ & R(2+1)D & $k=8$ & 77.8 & 49.1 \\

$\checkmark$ &$\checkmark$ & R(2+1)D & $k=8$ & \textbf{78.7} & \textbf{49.2} \\
$\checkmark$ &$\checkmark$ & R3D & $k=8$ & 74.1 & 43.6 \\
\bottomrule
\end{tabular}
\end{center}
\end{table}

\subsubsection{Temporally Adversarial Training}
We validate the effectiveness of temporally adversarial learning by showing the performance improvement upon the baseline. We first implement the baseline by not using the generator G during training. The standard memory queue from MoCo is also adopted. Second, we train the encoder by randomly dropping out several frames and use the remaining frames for adversarial learning. Third, we train the encoder by using the generator to adaptively drop out several frames. Besides training configurations, we analyze how the amount of dropout frames influences the downstream recognition performance. We choose to drop out $k$ frames ($k\in[4,8,12,16,24]$) and show the recognition performance accordingly.

Table~\ref{tab:adversararial} shows the analysis results. We notice that by using a random dropout scheme, the recognition performance degrades compared to the baseline performance. This is because random dropout does not enable VideoMoCo to learn temporally robust features. By introducing adversarial learning, the recognition performance is improved (i.e., $75.9\% / 46.2\%$ v.s $77.8\% / 49.1\%$) on both datasets. This indicates the effectiveness of our adversarial learning. Meanwhile, we analyze how the dropout amount influences the recognition performance by using different $k$. The results show that the highest performance is achieved when $k=8$. This is because a small $k$ (i.e., $k=4$) does not make the sample adversarial enough to train D, while a large $k$ (i.e., $k=[12,16,24]$) breaks the balance between G and D during adversarial learning. Furthermore, the performance of VideoMoCo is further improved by using temporal decay. This indicates that adversarial learning and temporal decay are effective to improve the baseline recognition performance.

\begin{table}
\begin{center}
\caption{Analysis on different values of temporal decay $t$. We experimentally find that $t=0.99999$ achieves the best.}
\label{tab:decay}
\vspace{1mm}
\renewcommand{\tabcolsep}{0.8mm}
\begin{tabular}{cccc}
\toprule
\multicolumn{3}{c}{Configuration} & Evaluation\\
\midrule
Adv. & Decay & Method & UCF101\\
\hline
$\checkmark$ &$\times$ & t=1 & 77.8\\
\midrule
$\checkmark$ &$\checkmark$ & $t=0.999$ & 73.4\\

$\checkmark$ &$\checkmark$ & $t=0.9999$ & 75.2 \\

$\checkmark$ &$\checkmark$ & $t=0.99999$ & 78.7\\

$\checkmark$ &$\checkmark$ & $t=0.999999$ & 78.3 \\

$\checkmark$ &$\checkmark$ & $t=0.9999999$ & 77.9 \\
\midrule
$\checkmark$ &$\checkmark$ & $t=0.99999$ & \textbf{78.7} \\
$\times$ &$\checkmark$ & $t=0.99999$ & 76.1\\
\bottomrule
\end{tabular}
\end{center}
\end{table}

\begin{table*}[t]
\begin{center}
\caption{Comparison with the state-of-the-art self-supervised learning methods on UCF101 and HMDB51 datasets. Our method VideoMoCo performs favorably against existing methods with a relatively small feature backbone.}
\label{tab:sota}
\vspace{2mm}
\renewcommand{\tabcolsep}{3mm}
\begin{tabular}{ccccccc}
\hline
\textbf{Method} & \textbf{Network} & \textbf{Input size} & \textbf{Params} & \textbf{Dataset} & \textbf{UCF101}  & \textbf{HMDB51}\\ \hline
Shuffle\&Learn~\cite{misra2016shuffle} & Alexnet& $256\times256$ & 62.4M & UCF101 & 50.2\% &18.1\% \\
Deep RL~\cite{buchler2018improving} & CaffeNet
& $227\times227$ & - & UCF101 & 58.6\% & 25.0\%  \\
OPN~\cite{lee2017unsupervised} & VGG-M-2048 & $80\times80$ & 8.6M & UCF101 & 59.8\% & 23.8\%\\
O3N~\cite{fernando2017self} & AlexNet & $227\times227$ & 62.4M & UCF101 & 60.3\% & 32.5\%\\
Spatio-Temp~\cite{wang2019self} & C3D &  $112\times112$  &  58.3M  & UCF101 & 58.8\% & 32.6\%\\
Spatio-Temp~\cite{wang2019self} & C3D & $112\times112$&  58.3M  & K-400 & 61.2\% & 33.4\%\\
VCOP~\cite{xu2019self}& C3D & $112\times112$ & 58.3M & UCF101 & 65.6\% & 28.4\%\\
RTT~\cite{jenni2020video}& C3D & $112\times112$ & 58.3M & UCF101 & 69.9\% & 39.6\%\\
DPC~\cite{han2019video} & 3D-ResNet34 & $224\times224$ & 32.6M & K-400 & 75.7\% & 35.7\%\\
\hline
RotNet3D~\cite{jing2018self} & 3D-ResNet18 & $224\times224$ & 33.6M & K-400 & 62.9\% & 33.7\% \\
ST-puzzle~\cite{kim2019self} & 3D-ResNet18 & $224\times224$ & 33.6M & K-400 & 65.8\% & 33.7\% \\
DPC~\cite{han2019video} & 3D-ResNet18 & $128\times128$ & 14.2M & K-400 & 68.2\% & 34.5\% \\
\textbf{Ours} & 3D-ResNet18 & $112\times112$ & 14.4M & K-400 & \textbf{74.1\%} & \textbf{43.6\%}\\
\hline
VCP~\cite{luo2020video}& R(2+1)D & $112\times112$ & 14.4M & UCF101 & 66.3\% & 32.2\%\\
VCOP~\cite{xu2019self} & R(2+1)D & $112\times112$ & 14.4M & UCF101 & 72.4\% & 30.9\%\\
PRP~\cite{yao2020video} & R(2+1)D & $112\times112$ & 14.4M & UCF101 & 72.1\% & 35.0\%\\
RTT~\cite{jenni2020video}& R(2+1)D & $112\times112$ & 14.4M & UCF101 & 81.6\% & 46.4\%\\
Pace Prediction~\cite{wang2020self} & R(2+1)D & $112\times112$ & 14.4M & K-400 & 77.1\% & 36.6\%\\
\textbf{Ours} & R(2+1)D & $112\times112$ & 14.4M & K-400 & \textbf{78.7\%} & \textbf{49.2\%}\\
\hline
\end{tabular}
\end{center}
\end{table*}

\subsubsection{Temporal Decay}
We analyze how temporal decay influences the recognition performance of VideoMoCo. We follow MoCo to set the size of the memory queue as 65536, and the momentum value of the encoder update is 0.999. The temporal decay $t$ we choose is $[0.999,0.9999,0.99999,0.999999,0.9999999]$. These values reflect how the increase of $t$ influences the recognition performance.

Table~\ref{tab:decay} shows the analysis results where $t=0.99999$ achieves the premier performance. When there is no temporal decay (i.e., $t=1$), keys in the queue contribute equally during contrastive learning. If $t$ is not relatively large (i.e., $t=0.999$), only the latest 8000 keys contribute to the learning process (i.e., $t^{8000}\approx0.0009$). To this end, we set $t=0.99999$ to ensure that the contributions are from all keys while the contribution from the oldest key halves (i.e., $t^{65536}\approx0.52$).

\subsection{Comparison with state-of-the-art Approaches}
We compare our approach with several state-of-the-art self-supervised video representation learning methods in both UCF101 and HMDB51 datasets. Table \ref{tab:sota} shows the evaluation results where the architectures, input size, number of parameters and pretrained dataset are illustrated as well.
The second block in Table \ref{tab:sota} indicates that our method performs favorably against existing approaches under the 3D-ResNet18 backbone on both UCF101 and HMDB51 datasets.
We notice that DPC uses R3D-34 as a feature backbone whose parameter amount is 3$\times$ more than ours. Also, the input size of DPC is larger than ours. Nevertheless, we achieve similar recognition performance on UCF101 and exceeds DPC on HMDB51. When using the same architecture (i.e., 3D-ResNet18) and similar input size, VideoMoCo outperforms DPC by a large margin (i.e., $5.9\%$ gain in UCF101 and $9.1\%$ gain in HMDB51).
We also evaluate our method using the R(2+1)D architecture and achieve the premier performance as shown in the third block of Table~\ref{tab:sota}. Specifically, VideoMoCo improves the existing method~\cite{wang2020self} under the same configuration. We note that even though VideoMoCo is trained using K-400 while other methods are trained using UCF101, the performance of VideoMoCo is still premier on both UCF101 and HMDB51 test sets. These evaluations indicate the favorable performance of VideoMoCo.

\section{Concluding Remarks}
We propose VideoMoCo for self-supervised video representation learning. Different from empirical pretext task investigation, we delve into MoCo and empower its temporal representation by introducing temporally adversarial learning and temporal decay. We treat the encoder as the discriminator and use a generator to perform adversarial learning. The generator augments training samples to robustify discriminators to capture temporally robust feature representations. The training process of the discriminator is from contrastive learning. Meanwhile, we propose temporal decay to model the attenuation of older keys in the queue. These keys ought to contribute less to the current input sample during the learning process. Our adversarial learning improves the temporal robustness of contrastive learning and the learned feature backbone is effective for downstream recognition tasks. The extensive experiments on the standard action recognition datasets UCF101 and HMDB50 demonstrate that our VideoMoCo has sufficiently improved MoCo and performs favorably against state-of-the-art self-supervised video representation learning approaches.

\clearpage

{\small
\bibliographystyle{ieee_fullname}
\bibliography{egbib}

\begin{thebibliography}{10}\itemsep=-1pt

\bibitem{buchler2018improving}
Uta Buchler, Biagio Brattoli, and Bjorn Ommer.
\newblock Improving spatiotemporal self-supervision by deep reinforcement
  learning.
\newblock In {\em European Conference on Computer Vision}, 2018.

\bibitem{chen2020simple}
Ting Chen, Simon Kornblith, Mohammad Norouzi, and Geoffrey Hinton.
\newblock A simple framework for contrastive learning of visual
  representations.
\newblock In {\em International Conference on Machine Learning}, 2020.

\bibitem{dwibedi2019temporal}
Debidatta Dwibedi, Yusuf Aytar, Jonathan Tompson, Pierre Sermanet, and Andrew
  Zisserman.
\newblock Temporal cycle-consistency learning.
\newblock In {\em IEEE/CVF Conference on Computer Vision and Pattern
  Recognition}, 2019.

\bibitem{fernando2017self}
Basura Fernando, Hakan Bilen, Efstratios Gavves, and Stephen Gould.
\newblock Self-supervised video representation learning with odd-one-out
  networks.
\newblock In {\em IEEE/CVF Conference on Computer Vision and Pattern
  Recognition}, 2017.

\bibitem{gan2018geometry}
Chuang Gan, Boqing Gong, Kun Liu, Hao Su, and Leonidas~J Guibas.
\newblock Geometry guided convolutional neural networks for self-supervised
  video representation learning.
\newblock In {\em IEEE/CVF Conference on Computer Vision and Pattern
  Recognition}, 2018.

\bibitem{goodfellow2014generative}
Ian Goodfellow, Jean Pouget-Abadie, Mehdi Mirza, Bing Xu, David Warde-Farley,
  Sherjil Ozair, Aaron Courville, and Yoshua Bengio.
\newblock Generative adversarial nets.
\newblock In {\em Neural Information Processing Systems}, 2014.

\bibitem{grill2020bootstrap}
Jean-Bastien Grill, Florian Strub, Florent Altch{\'e}, Corentin Tallec, Pierre
  Richemond, Elena Buchatskaya, Carl Doersch, Bernardo Avila~Pires, Zhaohan
  Guo, Mohammad Gheshlaghi~Azar, et~al.
\newblock Bootstrap your own latent-a new approach to self-supervised learning.
\newblock {\em Neural Information Processing Systems}, 2020.

\bibitem{gurumurthy2017deligan}
Swaminathan Gurumurthy, Ravi Kiran~Sarvadevabhatla, and R Venkatesh~Babu.
\newblock Deligan: Generative adversarial networks for diverse and limited
  data.
\newblock In {\em IEEE/CVF Conference on Computer Vision and Pattern
  Recognition}, 2017.

\bibitem{hadsell2006dimensionality}
Raia Hadsell, Sumit Chopra, and Yann LeCun.
\newblock Dimensionality reduction by learning an invariant mapping.
\newblock In {\em IEEE/CVF Conference on Computer Vision and Pattern
  Recognition}, 2006.

\bibitem{han2019video}
Tengda Han, Weidi Xie, and Andrew Zisserman.
\newblock Video representation learning by dense predictive coding.
\newblock In {\em Proceedings of the IEEE International Conference on Computer
  Vision Workshops}, 2019.

\bibitem{he2020momentum}
Kaiming He, Haoqi Fan, Yuxin Wu, Saining Xie, and Ross Girshick.
\newblock Momentum contrast for unsupervised visual representation learning.
\newblock In {\em IEEE/CVF Conference on Computer Vision and Pattern
  Recognition}, 2020.

\bibitem{he2019bag}
Tong He, Zhi Zhang, Hang Zhang, Zhongyue Zhang, Junyuan Xie, and Mu Li.
\newblock Bag of tricks for image classification with convolutional neural
  networks.
\newblock In {\em IEEE/CVF Conference on Computer Vision and Pattern
  Recognition}, 2019.

\bibitem{isola2017image}
Phillip Isola, Jun-Yan Zhu, Tinghui Zhou, and Alexei~A Efros.
\newblock Image-to-image translation with conditional adversarial networks.
\newblock In {\em IEEE/CVF Conference on Computer Vision and Pattern
  Recognition}, 2017.

\bibitem{jenni2020video}
Simon Jenni, Givi Meishvili, and Paolo Favaro.
\newblock Video representation learning by recognizing temporal
  transformations.
\newblock In {\em European Conference on Computer Vision}, 2020.

\bibitem{jhuang2011large}
H Jhuang, H Garrote, E Poggio, T Serre, and T Hmdb.
\newblock A large video database for human motion recognition.
\newblock In {\em IEEE/CVF International Conference on Computer Vision}, 2011.

\bibitem{jia2020robust}
Shuai Jia, Chao Ma, Yibing Song, and Xiaokang Yang.
\newblock Robust tracking against adversarial attacks.
\newblock In {\em European Conference on Computer Vision}, 2020.

\bibitem{jing2018self}
Longlong Jing, Xiaodong Yang, Jingen Liu, and Yingli Tian.
\newblock Self-supervised spatiotemporal feature learning via video rotation
  prediction.
\newblock {\em arXiv preprint arXiv:1811.11387}, 2018.

\bibitem{kay2017kinetics}
Will Kay, Joao Carreira, Karen Simonyan, Brian Zhang, Chloe Hillier, Sudheendra
  Vijayanarasimhan, Fabio Viola, Tim Green, Trevor Back, Paul Natsev, et~al.
\newblock The kinetics human action video dataset.
\newblock {\em arXiv preprint arXiv:1705.06950}.

\bibitem{kim2019self}
Dahun Kim, Donghyeon Cho, and In~So Kweon.
\newblock Self-supervised video representation learning with space-time cubic
  puzzles.
\newblock In {\em Proceedings of the AAAI Conference on Artificial
  Intelligence}, 2019.

\bibitem{kolesnikov2019big}
Alexander Kolesnikov, Lucas Beyer, Xiaohua Zhai, Joan Puigcerver, Jessica Yung,
  Sylvain Gelly, and Neil Houlsby.
\newblock Big transfer (bit): General visual representation learning.
\newblock In {\em European Conference on Computer Vision}, 2020.

\bibitem{kolesnikov2019revisiting}
Alexander Kolesnikov, Xiaohua Zhai, and Lucas Beyer.
\newblock Revisiting self-supervised visual representation learning.
\newblock In {\em IEEE/CVF Conference on Computer Vision and Pattern
  Recognition}, 2019.

\bibitem{korbar2018cooperative}
Bruno Korbar, Du Tran, and Lorenzo Torresani.
\newblock Cooperative learning of audio and video models from self-supervised
  synchronization.
\newblock In {\em Neural Information Processing Systems}, 2018.

\bibitem{lee2017unsupervised}
Hsin-Ying Lee, Jia-Bin Huang, Maneesh Singh, and Ming-Hsuan Yang.
\newblock Unsupervised representation learning by sorting sequences.
\newblock In {\em IEEE/CVF International Conference on Computer Vision}, 2017.

\bibitem{linsker1988self}
Ralph Linsker.
\newblock Self-organization in a perceptual network.
\newblock {\em Computer}, 21(3):105--117, 1988.

\bibitem{liu2020rethinking}
Hongyu Liu, Bin Jiang, Yibing Song, Wei Huang, and Chao Yang.
\newblock Rethinking image inpainting via a mutual encoder-decoder with feature
  equalizations.
\newblock In {\em European Conference on Computer Vision}, 2020.

\bibitem{luo2020video}
Dezhao Luo, Chang Liu, Yu Zhou, Dongbao Yang, Can Ma, Qixiang Ye, and Weiping
  Wang.
\newblock Video cloze procedure for self-supervised spatio-temporal learning.
\newblock {\em arXiv preprint arXiv:2001.00294}, 2020.

\bibitem{martin2017wasserstein}
SC Martin~Arjovsky and Leon Bottou.
\newblock Wasserstein generative adversarial networks.
\newblock In {\em International Conference on Machine Learning}, 2017.

\bibitem{misra2020self}
Ishan Misra and Laurens van~der Maaten.
\newblock Self-supervised learning of pretext-invariant representations.
\newblock In {\em IEEE/CVF Conference on Computer Vision and Pattern
  Recognition}, 2020.

\bibitem{misra2016shuffle}
Ishan Misra, C~Lawrence Zitnick, and Martial Hebert.
\newblock Shuffle and learn: unsupervised learning using temporal order
  verification.
\newblock In {\em European Conference on Computer Vision}, 2016.

\bibitem{nowozin2016f}
Sebastian Nowozin, Botond Cseke, and Ryota Tomioka.
\newblock f-gan: Training generative neural samplers using variational
  divergence minimization.
\newblock In {\em Neural Information Processing Systems}, 2016.

\bibitem{oord2018representation}
Aaron van~den Oord, Yazhe Li, and Oriol Vinyals.
\newblock Representation learning with contrastive predictive coding.
\newblock {\em arXiv preprint arXiv:1807.03748}, 2018.

\bibitem{poole2019variational}
Ben Poole, Sherjil Ozair, Aaron van~den Oord, Alexander~A Alemi, and George
  Tucker.
\newblock On variational bounds of mutual information.
\newblock 2019.

\bibitem{sabokrou2019self}
Mohammad Sabokrou, Mohammad Khalooei, and Ehsan Adeli.
\newblock Self-supervised representation learning via neighborhood-relational
  encoding.
\newblock In {\em IEEE/CVF International Conference on Computer Vision}, 2019.

\bibitem{song2017crest}
Yibing Song, Chao Ma, Lijun Gong, Jiawei Zhang, Rynson~WH Lau, and Ming-Hsuan
  Yang.
\newblock Crest: Convolutional residual learning for visual tracking.
\newblock In {\em IEEE/CVF International Conference on Computer Vision}, 2017.

\bibitem{song2018vital}
Yibing Song, Chao Ma, Xiaohe Wu, Lijun Gong, Linchao Bao, Wangmeng Zuo, Chunhua
  Shen, Rynson~WH Lau, and Ming-Hsuan Yang.
\newblock Vital: Visual tracking via adversarial learning.
\newblock In {\em IEEE/CVF Conference on Computer Vision and Pattern
  Recognition}, 2018.

\bibitem{soomro2012dataset}
Khurram Soomro, Amir~Roshan Zamir, and M Shah.
\newblock A dataset of 101 human action classes from videos in the wild.
\newblock {\em Center for Research in Computer Vision}, 2012.

\bibitem{souly2017semi}
Nasim Souly, Concetto Spampinato, and Mubarak Shah.
\newblock Semi supervised semantic segmentation using generative adversarial
  network.
\newblock In {\em IEEE/CVF International Conference on Computer Vision}, 2017.

\bibitem{srivastava2015unsupervised}
Nitish Srivastava, Elman Mansimov, and Ruslan Salakhudinov.
\newblock Unsupervised learning of video representations using lstms.
\newblock In {\em International Conference on Machine Learning}, 2015.

\bibitem{sun2019videobert}
Chen Sun, Austin Myers, Carl Vondrick, Kevin Murphy, and Cordelia Schmid.
\newblock Videobert: A joint model for video and language representation
  learning.
\newblock In {\em IEEE/CVF International Conference on Computer Vision}, 2019.

\bibitem{tian2020contrastive}
Yonglong Tian, Dilip Krishnan, and Phillip Isola.
\newblock Contrastive multiview coding.
\newblock In {\em European Conference on Computer Vision}, 2020.

\bibitem{vondrick2018tracking}
Carl Vondrick, Abhinav Shrivastava, Alireza Fathi, Sergio Guadarrama, and Kevin
  Murphy.
\newblock Tracking emerges by colorizing videos.
\newblock In {\em European Conference on Computer Vision}, 2018.

\bibitem{wang2019self}
Jiangliu Wang, Jianbo Jiao, Linchao Bao, Shengfeng He, Yunhui Liu, and Wei Liu.
\newblock Self-supervised spatio-temporal representation learning for videos by
  predicting motion and appearance statistics.
\newblock In {\em IEEE/CVF Conference on Computer Vision and Pattern
  Recognition}, 2019.

\bibitem{wang2020self}
Jiangliu Wang, Jianbo Jiao, and Yun-Hui Liu.
\newblock Self-supervised video representation learning by pace prediction.
\newblock In {\em European Conference on Computer Vision}, 2020.

\bibitem{wang2019unsupervised}
Ning Wang, Yibing Song, Chao Ma, Wengang Zhou, Wei Liu, and Houqiang Li.
\newblock Unsupervised deep tracking.
\newblock In {\em IEEE/CVF Conference on Computer Vision and Pattern
  Recognition}, 2019.

\bibitem{wang2020unsupervised}
Ning Wang, Wengang Zhou, Yibing Song, Chao Ma, Wei Liu, and Houqiang Li.
\newblock Unsupervised deep representation learning for real-time tracking.
\newblock {\em International Journal of Computer Vision}, 2020.

\bibitem{wang2019learning}
Xiaolong Wang, Allan Jabri, and Alexei~A Efros.
\newblock Learning correspondence from the cycle-consistency of time.
\newblock In {\em IEEE/CVF Conference on Computer Vision and Pattern
  Recognition}, 2019.

\bibitem{wang2020rethinking}
Yinglong Wang, Yibing Song, Chao Ma, and Bing Zeng.
\newblock Rethinking image deraining via rain streaks and vapors.
\newblock In {\em European Conference on Computer Vision}, 2020.

\bibitem{wu2018unsupervised}
Zhirong Wu, Yuanjun Xiong, Stella~X Yu, and Dahua Lin.
\newblock Unsupervised feature learning via non-parametric instance
  discrimination.
\newblock In {\em IEEE/CVF Conference on Computer Vision and Pattern
  Recognition}, 2018.

\bibitem{xingjian2015convolutional}
Shi Xingjian, Zhourong Chen, Hao Wang, Dit-Yan Yeung, Wai-Kin Wong, and
  Wang-chun Woo.
\newblock Convolutional lstm network: A machine learning approach for
  precipitation nowcasting.
\newblock In {\em Neural Information Processing Systems}, 2015.

\bibitem{xu2019self}
Dejing Xu, Jun Xiao, Zhou Zhao, Jian Shao, Di Xie, and Yueting Zhuang.
\newblock Self-supervised spatiotemporal learning via video clip order
  prediction.
\newblock In {\em IEEE/CVF Conference on Computer Vision and Pattern
  Recognition}, 2019.

\bibitem{yang2020video}
Ceyuan Yang, Yinghao Xu, Bo Dai, and Bolei Zhou.
\newblock Video representation learning with visual tempo consistency.
\newblock {\em arXiv preprint arXiv:2006.15489}, 2020.

\bibitem{yao2020video}
Yuan Yao, Chang Liu, Dezhao Luo, Yu Zhou, and Qixiang Ye.
\newblock Video playback rate perception for self-supervised spatio-temporal
  representation learning.
\newblock In {\em IEEE/CVF Conference on Computer Vision and Pattern
  Recognition}, 2020.

\bibitem{zhan2020online}
Xiaohang Zhan, Jiahao Xie, Ziwei Liu, Yew-Soon Ong, and Chen~Change Loy.
\newblock Online deep clustering for unsupervised representation learning.
\newblock In {\em IEEE/CVF Conference on Computer Vision and Pattern
  Recognition}, 2020.

\end{thebibliography}
}

\end{document}